\documentclass{sig-alternate}
\newfont{\mycrnotice}{ptmr8t at 7pt}
\newfont{\myconfname}{ptmri8t at 7pt}
\let\crnotice\mycrnotice%
\let\confname\myconfname%

\usepackage[german,american]{babel}
\usepackage[T1]{fontenc}
\usepackage[utf8]{inputenc}
\usepackage{times}
\usepackage{textcomp}
\usepackage{graphicx}
\usepackage{amssymb}
\usepackage{amsmath}
\usepackage{microtype} 

\usepackage{url}
\usepackage{hyperref}
\usepackage{multirow}
\usepackage{hhline}
\newcommand*\concat{\mathbin{\|}}
\usepackage{bm}
\usepackage{flushend}

\renewcommand{\tt}{\fontencoding{OT1}\fontfamily{cmtt}\selectfont}

\usepackage[usenames,dvipsnames]{color}
\usepackage{booktabs}
\usepackage[numbers,sectionbib]{natbib}
\makeatletter
\def\@copyrightspace{\relax}
\makeatother

\begin{document}

\title{Clickbait Detection in Tweets Using Self-attentive Network}
\subtitle{The Zingel Clickbait Detector at the Clickbait Challenge 2017}

\numberofauthors{1}
\author{
\alignauthor
Yiwei Zhou\thanks{Work performed while at The Alan Turing Institute.}\\
\affaddr{Department of Computer Science, University of Warwick, UK}\\
\affaddr{Yiwei.Zhou@warwick.ac.uk}\\
}

\maketitle

\begin{abstract}
Clickbait detection in tweets remains an elusive challenge.
In this paper, we describe the solution for the Zingel Clickbait Detector at the Clickbait Challenge 2017, which is capable of evaluating each tweet's level of click baiting.
We first reformat the regression problem as a multi-classification problem, based on the annotation scheme.
To perform multi-classification, we apply a token-level, self-attentive mechanism on the hidden states of bi-directional Gated Recurrent Units (biGRU), which enables the model to generate tweets' task-specific vector representations by attending to important tokens.
The self-attentive neural network can be trained end-to-end, without involving any manual feature engineering.
Our detector ranked $1^{st}$ in the final evaluation of Clickbait Challenge 2017.
\end{abstract}

\section{Introduction}

Clickbait refers to something (such as a headline) designed to make readers want to click on a hyperlink especially when the link leads to content of dubious value or interest\footnote{\url{https://www.merriam-webster.com/dictionary/clickbait}}.
With the rise of Twitter, publishers have been adopting various techniques to create the Curiosity Gap \cite{loewenstein1994psychology} between the information contained in the posted texts and the information the readers really want to know.
This gap drives the readers to click on the links contained in the tweets, and visit the publishers' websites.
According to \cite{potthast:2016}, all the top 20 most prolific publishers on Twitter employed clickbait on a regular basis, and the percentage of clickbait tweets reached an astonishing 26\% among all the tweets they published.

The clickbait tweets deliberately omit importance information, as well as include exaggerating and misleading information in the posted texts, as can be seen from the following examples:
\begin{itemize}
\item If you've ever used Google Docs for anything important, you should know about this.
\item This Tumblr account will nail your personality down in a second.
\item Here comes (almost) free money.
\end{itemize}

Clickbait draws negative impacts on readers, publishers, as well as social media websites.
First, it wastes the readers' time, leaves them feel disappointed and annoyed.
Second, it damages the publishers' reputation, as it violates the general codes of ethics of journalism.
Third, the traffic of the social media websites will be negatively affected if clogged up with low-quality and formulaic click baiting content.

This paper seeks to employ automatic approach to filter out clickbait tweets in the tweet stream.
We present the first attempt of applying self-attentive neural network to evaluate each tweet's level of click baiting, the effectiveness and efficiency of which has been proven in the Clickbait Challenge 2017 \cite{potthast:2017a}.

\section{Related Work}
Researchers have been exploring automatic approaches to perform Clickbait Detection.
However, most of the attempts focused on news headlines.
In \cite{chen2015misleading}, researchers suggested that ``clickbait can be identified through a
consideration of the existence of certain linguistic patterns, such
as the use of suspenseful language, unresolved pronouns, a
reversal narrative style, forward referencing, image placement,
reader’s behaviour and other important cues.''
However, their did not validate their conjecture by constructing corresponding automatic clickbait detector. 
Researchers in \cite{biyani20168} extracted content features,  textual similarity features between the headline and the body, as well as informality and forward reference features from the news on Yahoo homepage, and trained Gradient Boosted Decision Trees to decide if a news article was a clickbait.
Feature engineering was also employed in \cite{chakraborty2016stop}, where researchers constructed vector representations for Wikinews headlines by extracting sentence structure features, word pattern features, clickbait language features, as well as N-gram features.
They further trained a SVM classifier to perform binary classification on the resulting vector representations.
Their results were further improved by researchers in \cite{anand2017we} and \cite{rony2017diving}, by using bidirectional Recurrent Neural Network (RNN) \cite{schuster1997bidirectional} and fastText \cite{joulin2016bag} on word distributed representations, respectively.
Following \cite{biyani20168,chakraborty2016stop}, researchers in \cite{wei2017learning} went back to manual feature engineering, to extract various body dependent and body independent features from Chinese news articles; then they applied the co-training algorithm to make use of the unlabelled data set.
Besides the texts, researchers in \cite{zheng2017boost} additionally took the user behaviour information into consideration, to improve the performance of clickbait detection on Chinese news articles.

There are some recent researches targeting at training clickbait detectors on posts from social media websites, such as Twitter.
For example, researchers in \cite{potthast:2016} trained a random forest classifier by extracting various features from the post texts, linked webpages and associated meta information of tweets, to decide if a tweet was a clickbait.
Researchers in \cite{agrawal2016clickbait} trained a Convolutional Neural Network (CNN) \cite{kim2014convolutional}, using the post texts only, to detect clickbait posts in Reddit, Facebook and Twitter.
In \cite{chakraborty2018tabloids}, researchers analysed the differences in content, sentiment, consumers, etc., between the clickbait tweets and non-clickbait tweets. 

All of former researches treat automatic clickbait detection as a binary classification problem.
According to \cite{potthast:2017b}, whether a tweet was a clickbait was not a simple black and white problem, thus it was better to associate each tweet with a graded value reflecting its level of click baiting, rather than a binary indicator.
Thus, different from former researches, this paper tackles the Clickbait Challenge 2017\footnote{\url{http://www.clickbait-challenge.org/}} \cite{potthast:2017a}, i.e., rating how click baiting each tweet is.

Since proposed in \cite{bahdanau2014neural}, the attention mechanism has been applied on various text classification-related problems.
For example, aspect-based sentiment analysis \cite{ruder2016hierarchical,wang2016attention}, target-specific stance detection \cite{zhou2017attention}, and fake news detection \cite{chopra2017towards}.
The attention mechanism releases the text encoders from the burden of encoding all the information in the text into one fixed length vector, and allows the neural network model to automatically attend to tokens that are important for the prediction.
The above researches leveraged extra information to form a context vector, which was further employed to infer the weights of token embeddings in the text, when aggregating them into one task-specific sentence embedding.
However, for the Clickbait Challenge 2017, there is no external information to employ, thus the context vector can only be learned from the text itself.
In this paper, we present the first attempt to apply the state-of-the-art self-attentive network \cite{yang2016hierarchical,lin2017structured} on Clickbait Detection in tweets.


\section{Approach}

%
\subsection{Problem Definition}
\label{sec:pd}
The Clickbait Challenge 2017 was defined as a regression problem, asking the competition teams to automatically evaluate the clickbait score of each tweet in the test dataset, indicating the its level of click baiting.
Instead of performing regression directly, we performed a detailed analysis of the annotation scheme to understand how the clickbait score in the training datasets was generated, and reformatted the problem.
When crowdsourcing the dataset, for each tweet, four categories were provided to five annotators, which were ``not click baiting'', ``slightly click baiting'', ``considerably click baiting'' and ``heavily click baiting'' \cite{potthast:2016,potthast:2017b}.
The above four categories were assigned with values 0, $\frac{1}{3}$, $\frac{2}{3}$ and 1.0, respectively, to demonstrate the tweet's level of click baiting.
As a result, each annotated tweet in the training datasets were associated with five fields, which were ``truthJudgements'', ``truthMean'', ``truthMedian'', ``truthMode'' and ``truthClass''.
The ``truthJudgements'' fields demonstrated the choices made by the five different annotators;
the ``truthMean'', ``truthMedian'' and ``truthMode'' fields were the mean, median and mode of the five annotators' choices, respectively, after mapping the choices into their corresponding values.
For one tweet, if the number of annotators saw it as ``not click baiting'' or ``slightly click baiting'', was larger than the number of annotators saw it as ``considerably click baiting'' and ``heavily click baiting'', then the ``truthClass'' of this tweet would be ``clickbait''. Otherwise, its ``truthClass'' would be ``no-clickbait''.

The clickbait detectors were asked to automatically infer the ``truthMean'' for each tweet in the test dataset.
Our Zingel Clickbait Detector reproduced the above process by generating the predicted annotation distribution $[p_1, p_2, p_3, p_4]$ for each tweet on $[0, \frac{1}{3}, \frac{2}{3}, 1]$, where $p_1+p_2+p_3+p_4=1$.
Based on the annotation distribution, the following calculations were performed:
\begin{equation}
truthMean  = p_1 \times 0 + p_2 \times \frac{1}{3} + p_3 \times \frac{2}{3} + p_4 \times 1.0;
\end{equation}
\begin{equation}
truthClass = \begin{cases}
                \text{non-clickbait}, & \text{if}\ p_1 + p_2 > p_3 + p_4
                \\
                \text{clickbait}, & \text{otherwise}.
               \end{cases}
\end{equation}
Similar strategy was applied by \cite{tai2015improved} to infer the semantic relatedness of sentence pairs.
\subsection{Self-attentive Network}
We applied the self-attentive network \cite{yang2016hierarchical,lin2017structured}, to tackle the annotation distribution prediction problem.
Self-attentive RNN applied a token level attention mechanism over the hidden states generated by the RNN, to infer the levels of importance of the tweet tokens in predicting the annotation distribution.
Different from regular external-attentive network \cite{bahdanau2014neural,ruder2016hierarchical,wang2016attention,zhou2017attention,chopra2017towards}, self-attentive network does not need any external information to learn the context vector.
The inferred information can be used as the tokens' weights, when aggregating their hidden states into the vector representation of the tweet.
\begin{figure}[htb]
	\centering
	\includegraphics[width=0.55\textwidth]{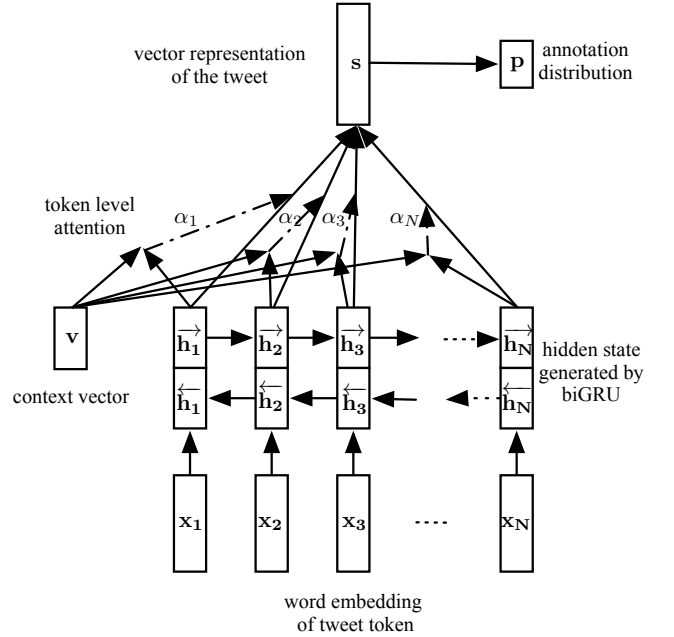}
	\caption{Self-attentive Network for Annotation Distribution Prediction.}
	\label{fig:1}
\end{figure}
Figure~\ref{fig:1} demonstrates the architecture of the self-attentive network.
Given a tweet whose ``postText'' contains $N$ tokens, we first maped each token $w_n$, where $n \in [1, N]$, to its corresponding word embedding $\mathbf{x_n}$, through an word embedding matrix $\mathbf{W_E} \in \mathbb{R}^{V \times d_0}$, where $d_0$ denotes the dimensionality of the word embedding, $V$ denotes the size of the vocabulary.
After that, we used a bi-directional GRU \cite{bahdanau2014neural} to encode the contextual information from both directions of the token into its hidden state:
\begin{equation}
\overrightarrow{\mathbf{h_n}} = \overrightarrow{GRU}(\mathbf{x_n}, \overrightarrow{\mathbf{h_{n-1}}}),
\end{equation}
\begin{equation}
\overleftarrow{\mathbf{h_n}} = \overleftarrow{GRU}(\mathbf{x_n}, \overleftarrow{\mathbf{h_{n+1}}}),
\end{equation}
where $\overrightarrow{\mathbf{h_n}}, \overleftarrow{\mathbf{h_n}} \in \mathbb{R}^{d_1}$.
The resulting hidden state of biGRU for each token was the concatenation of its forward hidden state and backward hidden states, i.e., 
\begin{equation}
\mathbf{h_n} = \overrightarrow{\mathbf{h_n}} \concat \overleftarrow{\mathbf{h_n}}.
\end{equation}
The token level attention vector $\bm{\alpha} \in \mathbb{R}^{N}$, which represents the weights of tokens in ``postText'' when predicting the annotation distribution, can be calculated as:
\begin{equation}
\bm{\alpha} = \text{softmax}(\tanh(\mathbf{H}\mathbf{W_H})\mathbf{v}),
\end{equation}
where $\mathbf{H} = [\mathbf{h_1}, \mathbf{h_2}, \cdots, \mathbf{h_N}]$, and $\mathbf{H} \in \mathbb{R}^{N \times 2d_1}$;
$\mathbf{W_H} \in \mathbb{R}^{2d_1 \times 2d_1}$, $\mathbf{v} \in \mathbb{R}^{2d_1}$ are the parameters to train; the $\text{softmax}(\cdot)$ function guarantees that $\sum_{n=1}^{N} \alpha_{n} = 1$.

The vector representation of the tweet's ``postText'' $\mathbf{s}$ can be calculated as:
\begin{equation}
\mathbf{s} = \mathbf{H}^{T}\bm{\alpha},
\end{equation}
where $\mathbf{s} \in \mathbb{R}^{2d_1}$.

The predicted annotation distribution of this tweet on four categories $\mathbf{p}=[p_1, p_2, p_3, p_4]$ can be calculated as:
\begin{equation}
\mathbf{p} = \text{softmax}(\mathbf{W_s}\mathbf{s}+\mathbf{b_s}),
\end{equation}
where $\mathbf{W_s} \in \mathbb{R}^{4 \times 2d_1}$ and the bias term $\mathbf{b_s} \in \mathbb{R}^{4}$ are the parameters to train.

Only the ``postText'' field of the each labelled sample was employed in our approach.
Because the human annotators made the judgements only based on the tweet’s plain text and image \cite{potthast:2016}. 
As a large proportion of tweets were not associated with images, we did not employ images when training the detector, so that the same detector can be applied to annotated all the tweets.
\section{Experiments and Results}
\subsection{Dataset Description}
In this challenge, three datasets were provided to the competition teams, which were Dataset A, Dataset B (Unlabelled), and Dataset C.
The statistics of these datasets are shown in Table~\ref{tab:1}.

\begin{table}[htbp]
\centering
\caption{Statistics of the datasets.}
\begin{tabular}{|c|c|c|c|} \hline
\textbf{Dataset} & \textbf{\# tweets} & \textbf{\# clickbait} & \textbf{\# no-clickbait}\\ \hline
\textbf{A} & 2,495 & 762 & 1,697 \\ \hline
\textbf{B} & 80,012 & N/A & N/A \\ \hline
\textbf{C} & 19,538 & 4,761 & 14,777  \\ \hline
\end{tabular}
\label{tab:1}
\end{table}
\subsection{Model Training}
We inferred the annotation distribution $\mathbf{p}$ on four categories from the ``truthJudgments'' field, for all the labelled samples.
The self-attentive network was trained by optimising the cross-entropy loss between the true annotation distribution and the predicted annotation distribution.
We used the Adam optimiser \cite{kingma2014adam} to train the parameters in the self-attentive network, which had the best overall performances on sparse training data, according to \cite{ruder2016overview}.

To make the trained models less prone to overfitting, we applied dropout on the outputs of the word embedding layer, on the outputs of the biGRU encoding layer, as well as on the outputs of the self-attentive layer \cite{srivastava2014dropout,pham2014dropout}. 
To reduce the space of hyper-parameter searching, we set $d_0$ to 100, set $d_1$ to 64, set a dropout rate of 0.2 on the outputs of the word embedding layer, set a dropout rate of 0.3 on the outputs of the biGRU encoding layer, based on our former neural network training experiences; because we observed these hyper-parameters had very limited influence on the performance of the clickbait detector. 
We initialised the word embedding matrix $\mathbf{W_E}$ with the 100-dimension pre-trained Glove embeddings \cite{pennington2014glove} of Wikipedia data.
We performed additional experiments by initialising $\mathbf{W_E}$ with 100-dimension pre-trained Glove embeddings of tweets, as well as 100-dimension pre-trained Word2Vec embeddings \cite{mikolov2013efficient} of Data B, but their performances were not comparable with our initial choice.
The word embedding matrix $\mathbf{W_E}$ was updated during the training process, to be adjusted for the clickbait detection scenario.

The same preprocessing step was adopted as the one applied by Pennington et al. \cite{pennington2014glove} to generate pre-trained embeddings of tweets\footnote{\url{https://nlp.stanford.edu/projects/glove/preprocess-twitter.rb}}.
The preprocessed tweets was tokenised by the TweetTokenizer in NLTK\footnote{\url{http://www.nltk.org/}}.
We used the maximum of ``postText'' token numbers in the training dataset(s) as the input length $N$, thus all the ``postText''s of tweets were either zero-padded or truncated to the same length.
All the tokens that were not included in the pre-trained Glove embeddings of Wikipedia data were replaced by ``<unk>'', the embedding of which was initialised by random numbers between $-0.1$ and $0.1$.

For the rest hyper-parameters, we performed more rigid hyper-parameter optimisation, using Hyperopt \cite{bergstra2013hyperopt}, through 5-fold cross-validation on Dataset C only.
Specifically, we selected the batch size from $[16, \mathbf{32}, 64, 128]$; we selected the dropout rate on the outputs of the self-attentive layer from $[0.3, \mathbf{0.5}, 0.7]$; we selected the learning rate of the Adam optimiser from $[0.001, \mathbf{0.005}, 0.01, 0.05]$; we also applied gradient clipping \cite{pascanu2013difficulty} to avoid gradient explosion, and the clipping cutoff threshold was selected from $[0.5, 1, \mathbf{2}, 5, 10]$.
For each round of cross-validation, we trained the neural network for a maximum number of 20 epochs on 80\% samples from Dataset C ($\approx$ 15630 samples); and used the detector's Mean Squared Error (MSE) on the remaining samples from Dataset C to determine the number of epochs.
We averaged the least MSEs for all rounds of cross-validation for each hyper-parameter set, and used it as the criterion to select the best combination of hyper-parameters, which were demonstrated above as bold numbers among various choices.

We did not employ Dataset A during the hyper-parameter optimisation step, so that we can get an idea of the detector's performance on unseen data.

We implemented the self-attentive network using Tensorflow \cite{45166}, and the code is available at \url{https://github.com/zhouyiwei/cc}.
\subsection{Results}
In our final submission, we combined Dataset A and Dataset C together, to leverage all the labelled data for training.
Since the number of epochs was essential for the detector's performance, we applied the same strategy as the hyper-parameter optimisation step.
Specifically, we trained 5 self-attentive neural networks, with each using 80\% of the labelled samples from the combined dataset for training, and the remaining 20\% for validation to decide the number of epochs.
All the hyper-parameters of the 5 neural networks were the same, which were selected in the hyper-parameter optimisation step.
For each sample in the test dataset, the annotation distributions generated by the 5 neural networks were averaged to generate the final ``truthMean'' and ``truthClass'' prediction, as described in Section~\ref{sec:pd}.
The evaluation was performed on a Ubuntu 16.04 server provided by the TIRA platform \cite{potthast:2014}.
The Zingel Clickbait Detector achieved an MSE of 0.033 (ranked $1^{st}$) in the final evaluation, which substantially outperformed the baseline approach reported in \cite{potthast:2017a}, with an MSE of 0.044.
Besides that, our detector also outperformed all the other detectors in term of F1 score (0.683), Accuracy (0.856) and Running Time (00:03:27).
Thus, the Zingel Clickbait Detector achieved consistent and competitive performance across multiple evaluation metrics, with very low computational cost.
\section{Conclusion}
In this paper, we have presented the solution for the Zingel Clickbait Detector.
We tackled the clickbait score prediction problem in Clickbait Challenge 2017 by performing multi-classification using the self-attentive neural network.
The Zingel Clickbait Detector has achieved the state-of-the-art performance on the Twitter Clickbait Score Prediction dataset provided by Clickbait Challenge 2017, in terms of MSE, F1 score, Accuracy and Running time.

\section*{Acknowledgments}
This work was supported by The Alan Turing Institute under the EPSRC grant EP/N510129/1.

\begin{raggedright}
\bibliography{clickbait17-notebook-lit}

\begin{thebibliography}{34}
\providecommand{\natexlab}[1]{#1}
\providecommand{\url}[1]{\texttt{#1}}
\expandafter\ifx\csname urlstyle\endcsname\relax
  \providecommand{\doi}[1]{doi: #1}\else
  \providecommand{\doi}{doi: \begingroup \urlstyle{rm}\Url}\fi

\bibitem[Abadi et~al.(2015)Abadi, Agarwal, Barham, Brevdo, Chen, Citro,
  Corrado, Davis, Dean, Devin, et~al.]{45166}
M.~Abadi, A.~Agarwal, P.~Barham, E.~Brevdo, Z.~Chen, C.~Citro, G.~S. Corrado,
  A.~Davis, J.~Dean, M.~Devin, et~al.
\newblock Tensorflow: Large-scale machine learning on heterogeneous distributed
  systems, 2015.

\bibitem[Agrawal(2016)]{agrawal2016clickbait}
A.~Agrawal.
\newblock Clickbait detection using deep learning.
\newblock In \emph{Proceedings of the 2nd International Conference on Next
  Generation Computing Technologies}, pages 268--272. IEEE, 2016.

\bibitem[Anand et~al.(2017)Anand, Chakraborty, and Park]{anand2017we}
A.~Anand, T.~Chakraborty, and N.~Park.
\newblock We used neural networks to detect clickbaits: You won’t believe
  what happened next!
\newblock In \emph{Proceedings of the 39th European Conference on Information
  Retrieval}, pages 541--547. Springer, 2017.

\bibitem[Bahdanau et~al.(2015)Bahdanau, Cho, and Bengio]{bahdanau2014neural}
D.~Bahdanau, K.~Cho, and Y.~Bengio.
\newblock Neural machine translation by jointly learning to align and
  translate.
\newblock In \emph{Proceedings of the 3rd International Conference on Learning
  Representations}, 2015.

\bibitem[Bergstra et~al.(2013)Bergstra, Yamins, and Cox]{bergstra2013hyperopt}
J.~Bergstra, D.~Yamins, and D.~D. Cox.
\newblock Hyperopt: A python library for optimizing the hyperparameters of
  machine learning algorithms.
\newblock In \emph{Proceedings of the 12th Python in Science Conference}, pages
  13--20, 2013.

\bibitem[Biyani et~al.(2016)Biyani, Tsioutsiouliklis, and
  Blackmer]{biyani20168}
P.~Biyani, K.~Tsioutsiouliklis, and J.~Blackmer.
\newblock ``8 amazing secrets for getting more clicks'': Detecting clickbaits
  in news streams using article informality.
\newblock In \emph{Proceedings of the 30th AAAI Conference on Artificial
  Intelligence}, pages 94--100. AAAI, 2016.

\bibitem[Chakraborty et~al.(2016)Chakraborty, Paranjape, Kakarla, and
  Ganguly]{chakraborty2016stop}
A.~Chakraborty, B.~Paranjape, S.~Kakarla, and N.~Ganguly.
\newblock Stop clickbait: Detecting and preventing clickbaits in online news
  media.
\newblock In \emph{Proceedings of the 2016 International Conference on Advances
  in Social Networks Analysis and Mining}, pages 9--16. IEEE, 2016.

\bibitem[Chakraborty et~al.(2018)Chakraborty, Sarkar, Mrigen, and
  Ganguly]{chakraborty2018tabloids}
A.~Chakraborty, R.~Sarkar, A.~Mrigen, and N.~Ganguly.
\newblock Tabloids in the era of social media? understanding the production and
  consumption of clickbaits in twitter.
\newblock In \emph{Proceedings of the 21st ACM Conference on Computer-Supported
  Cooperative Work and Social Computing}. ACM, 2018.

\bibitem[Chen et~al.(2015)Chen, Conroy, and Rubin]{chen2015misleading}
Y.~Chen, N.~J. Conroy, and V.~L. Rubin.
\newblock Misleading online content: Recognizing clickbait as false news.
\newblock In \emph{Proceedings of the 2015 ACM Workshop on Multimodal Deception
  Detection}, pages 15--19. ACM, 2015.

\bibitem[Chopra et~al.(2017)Chopra, Jain, and Sholar]{chopra2017towards}
S.~Chopra, S.~Jain, and J.~M. Sholar.
\newblock Towards automatic identification of fake news: Headline-article
  stance detection with lstm attention models, 2017.

\bibitem[Grave et~al.(2017)Grave, Mikolov, Joulin, and
  Bojanowski]{joulin2016bag}
E.~Grave, T.~Mikolov, A.~Joulin, and P.~Bojanowski.
\newblock Bag of tricks for efficient text classification.
\newblock In \emph{Proceedings of the 15th Conference of the European Chapter
  of the Association for Computational Linguistics}, pages 427--431, 2017.

\bibitem[Kim(2014)]{kim2014convolutional}
Y.~Kim.
\newblock Convolutional neural networks for sentence classification.
\newblock In \emph{Proceedings of the 2014 Conference on Empirical Methods in
  Natural Language Processing}, pages 1746--1751. ACL, 2014.

\bibitem[Kingma and Ba(2015)]{kingma2014adam}
D.~Kingma and J.~Ba.
\newblock Adam: A method for stochastic optimization.
\newblock In \emph{Proceedings of the 3rd International Conference on Learning
  Representations : Poster Session}, 2015.

\bibitem[Lin et~al.(2017)Lin, Feng, Santos, Yu, Xiang, Zhou, and
  Bengio]{lin2017structured}
Z.~Lin, M.~Feng, C.~N.~d. Santos, M.~Yu, B.~Xiang, B.~Zhou, and Y.~Bengio.
\newblock A structured self-attentive sentence embedding.
\newblock In \emph{Proceedings of the 5th International Conference on Learning
  Representations}, 2017.

\bibitem[Loewenstein(1994)]{loewenstein1994psychology}
G.~Loewenstein.
\newblock The psychology of curiosity: A review and reinterpretation.
\newblock \emph{Psychological bulletin}, 116\penalty0 (1):\penalty0 75, 1994.

\bibitem[Mikolov et~al.(2013)Mikolov, Chen, Corrado, and
  Dean]{mikolov2013efficient}
T.~Mikolov, K.~Chen, G.~Corrado, and J.~Dean.
\newblock Efficient estimation of word representations in vector space.
\newblock In \emph{Proceedings of the 1st International Conference on Learning
  Representations : Workshop Track}, 2013.

\bibitem[Pascanu et~al.(2013)Pascanu, Mikolov, and
  Bengio]{pascanu2013difficulty}
R.~Pascanu, T.~Mikolov, and Y.~Bengio.
\newblock On the difficulty of training recurrent neural networks.
\newblock In \emph{Proceedings of the 30th International Conference on Machine
  Learning}, volume~28, pages 1310--1318, 2013.

\bibitem[Pennington et~al.(2014)Pennington, Socher, and
  Manning]{pennington2014glove}
J.~Pennington, R.~Socher, and C.~D. Manning.
\newblock Glove: Global vectors for word representation.
\newblock In \emph{Proceedings of the 2014 conference on Empirical Methods in
  Natural Language Processing}, volume~14, pages 1532--1543. ACL, 2014.

\bibitem[Pham et~al.(2014)Pham, Bluche, Kermorvant, and
  Louradour]{pham2014dropout}
V.~Pham, T.~Bluche, C.~Kermorvant, and J.~Louradour.
\newblock Dropout improves recurrent neural networks for handwriting
  recognition.
\newblock In \emph{Proceedings of the 14th International Conference on
  Frontiers in Handwriting Recognition}, pages 285--290, 2014.

\bibitem[Potthast et~al.(2014)Potthast, Gollub, Rangel, Rosso, Stamatatos, and
  Stein]{potthast:2014}
M.~Potthast, T.~Gollub, F.~Rangel, P.~Rosso, E.~Stamatatos, and B.~Stein.
\newblock {Improving the Reproducibility of PAN's Shared Tasks: Plagiarism
  Detection, Author Identification, and Author Profiling}.
\newblock In \emph{Proceedings of {CLEF} 2014 Conference and Labs of the
  Evaluation Forum}, pages 268--299. Springer, 2014.

\bibitem[Potthast et~al.(2016)Potthast, K{\"o}psel, Stein, and
  Hagen]{potthast:2016}
M.~Potthast, S.~K{\"o}psel, B.~Stein, and M.~Hagen.
\newblock {Clickbait Detection}.
\newblock In \emph{Proceedings of the 38th European Conference on Information
  Retrieval}, volume 9626, pages 810--817. Springer, 2016.

\bibitem[Potthast et~al.(2017{\natexlab{a}})Potthast, Gollub, Hagen, and
  Stein]{potthast:2017a}
M.~Potthast, T.~Gollub, M.~Hagen, and B.~Stein.
\newblock {The Clickbait Challenge 2017: Towards a Regression Model for
  Clickbait Strength}.
\newblock In \emph{{Proceedings of the Clickbait Challenge}},
  2017{\natexlab{a}}.

\bibitem[Potthast et~al.(2017{\natexlab{b}})Potthast, Gollub, Komlossy,
  Schuster, Wiegmann, Garces, Hagen, and Stein]{potthast:2017b}
M.~Potthast, T.~Gollub, K.~Komlossy, S.~Schuster, M.~Wiegmann, E.~Garces,
  M.~Hagen, and B.~Stein.
\newblock {Crowdsourcing a Large Corpus of Clickbait on Twitter}.
\newblock In \emph{{(to appear)}}, 2017{\natexlab{b}}.

\bibitem[Rony et~al.(2017)Rony, Hassan, and Yousuf]{rony2017diving}
M.~M.~U. Rony, N.~Hassan, and M.~Yousuf.
\newblock Diving deep into clickbaits: Who use them to what extents in which
  topics with what effects?
\newblock \emph{arXiv preprint arXiv:1703.09400}, 2017.

\bibitem[Ruder(2016)]{ruder2016overview}
S.~Ruder.
\newblock An overview of gradient descent optimization algorithms.
\newblock \emph{arXiv preprint arXiv:1609.04747}, 2016.

\bibitem[Ruder et~al.(2016)Ruder, Ghaffari, and Breslin]{ruder2016hierarchical}
S.~Ruder, P.~Ghaffari, and J.~G. Breslin.
\newblock A hierarchical model of reviews for aspect-based sentiment analysis.
\newblock In \emph{Proceedings of the 2016 Conference on Empirical Methods in
  Natural Language Processing}, pages 999--1005. ACL, 2016.

\bibitem[Schuster and Paliwal(1997)]{schuster1997bidirectional}
M.~Schuster and K.~K. Paliwal.
\newblock Bidirectional recurrent neural networks.
\newblock \emph{IEEE Transactions on Signal Processing}, 45\penalty0
  (11):\penalty0 2673--2681, 1997.

\bibitem[Srivastava et~al.(2014)Srivastava, Hinton, Krizhevsky, Sutskever, and
  Salakhutdinov]{srivastava2014dropout}
N.~Srivastava, G.~E. Hinton, A.~Krizhevsky, I.~Sutskever, and R.~Salakhutdinov.
\newblock Dropout: a simple way to prevent neural networks from overfitting.
\newblock \emph{Journal of Machine Learning Research}, 15\penalty0
  (1):\penalty0 1929--1958, 2014.

\bibitem[Tai et~al.(2015)Tai, Socher, and Manning]{tai2015improved}
K.~S. Tai, R.~Socher, and C.~D. Manning.
\newblock Improved semantic representations from tree-structured long
  short-term memory networks.
\newblock In \emph{Proceedings of the 53rd annual meeting of the Association
  for Computational Linguistics and the 7th International Joint Conference on
  Natural Language Processing of the Asian Federation of Natural Language
  Processing}, pages 1556--1566. ACL, 2015.

\bibitem[Wang et~al.(2016)Wang, Huang, Zhu, and Zhao]{wang2016attention}
Y.~Wang, M.~Huang, X.~Zhu, and L.~Zhao.
\newblock Attention-based lstm for aspect-level sentiment classification.
\newblock In \emph{Proceedings of the 2016 conference on Empirical Methods in
  Natural Language Processing}, pages 606--615. ACL, 2016.

\bibitem[Wei and Wan(2017)]{wei2017learning}
W.~Wei and X.~Wan.
\newblock Learning to identify ambiguous and misleading news headlines.
\newblock In \emph{Proceedings of the 26th International Joint Conference on
  Artificial Intelligence}, pages 4172--4178. AAAI, 2017.

\bibitem[Yang et~al.(2016)Yang, Yang, Dyer, He, Smola, and
  Hovy]{yang2016hierarchical}
Z.~Yang, D.~Yang, C.~Dyer, X.~He, A.~J. Smola, and E.~H. Hovy.
\newblock Hierarchical attention networks for document classification.
\newblock In \emph{Proceedings of the 2016 Conference of the North American
  Chapter of the Association for Computational Linguistics - Human Language
  Technologies}, pages 1480--1489. ACL, 2016.

\bibitem[Zheng et~al.(2017)Zheng, Yao, Jiang, Xia, and Xiao]{zheng2017boost}
H.-T. Zheng, X.~Yao, Y.~Jiang, S.-T. Xia, and X.~Xiao.
\newblock Boost clickbait detection based on user behavior analysis.
\newblock In \emph{Proceedings of the 1st Asia Pacific Web and Web-Age
  Information Management Joint Conference on Web and Big Data}, pages 73--80.
  Springer, 2017.

\bibitem[Zhou and Cristea(2017)]{zhou2017attention}
Y.~Zhou and A.~I. Cristea.
\newblock Connecting targets to tweets: Semantic attention-based model for
  target-specific stance detection.
\newblock In \emph{Proceedings of the 18th International Conference on Web
  Information Systems Engineering}, pages 18--32. Springer, 2017.

\end{thebibliography}
\end{raggedright}
\end{document}